\documentclass[10pt]{article} 
\usepackage{amsmath,amsfonts,amssymb,amsthm} %math stuff
\usepackage{mathtools} %math stuff
\usepackage{graphicx,float} %figures
\usepackage{algorithm, algpseudocode, algorithmicx} %psuedo code
\usepackage{bm} %bold greek letters
\usepackage{bbm} %indicator function
\usepackage{fullpage} %margins
\usepackage[top=1.9 cm, bottom = 1.9 cm, left = 1.9 cm, right = 1.9cm]{geometry} %margins
\usepackage[hidelinks]{hyperref} %\label and \ref
\hypersetup{
colorlinks = true,
urlcolor = blue,
linkcolor = red,
citecolor = green
}
\usepackage{enumitem} %for customizing itemize/enumerate environments
\usepackage{grffile} %allows you to name pictures with spaces in the name

\newcommand{\mathe}{\mathbb{E}}
\newcommand{\norm}[1]{\left\lVert\,#1\,\right\rVert}

\newcommand{\innerp}[1]{\left\langle#1 \right\rangle}
\newtheorem{theorem}{Theorem}

\title{Variance Reduction for Score Functions Using Optimal Baselines}
\author{\large Ronan Keane$^{1,3}$ \and H. Oliver Gao$^{1, 2}$}
\date{$^1$Systems Science and Engineering, Cornell University \\
$^2$Civil and Environmental Engineering, Cornell University \\
$^3$Email: (rlk268@cornell.edu) \\ \vspace{2mm}
July 28, 2022}

\begin{document}
\maketitle 
\section*{Abstract}
Many problems involve the use of models which learn probability distributions or incorporate randomness in some way. In such problems, because computing the true expected gradient may be intractable, a gradient estimator is used to update the model parameters. When the model parameters directly affect a probability distribution, the gradient estimator will involve score function terms. This paper studies baselines, a variance reduction technique for score functions. Motivated primarily by reinforcement learning, we derive for the first time an expression for the optimal state-dependent baseline, the baseline which results in a gradient estimator with minimum variance. Although we show that there exist examples where the optimal baseline may be arbitrarily better than a value function baseline, we find that the value function baseline usually performs similarly to an optimal baseline in terms of variance reduction. Moreover, the value function can also be used for bootstrapping estimators of the return, leading to additional variance reduction. Our results give new insight and justification for why value function baselines and the generalized advantage estimator (GAE) work well in practice. \\

\noindent \textit{Keywords:} score function, likelihood ratio method, baselines, reinforcement learning, control variates, generalized advantage estimator
\pagebreak

\section{Introduction}
It seems that in recent years, stochastic models have become increasingly popular and relevant to many different applications in science and engineering. When dealing with such models, we often rely on a gradient-based optimization algorithm such as stochastic gradient descent (SGD) to update the model parameters. One notable aspect of SGD is that it does not use the true expected gradient, instead using a gradient estimator, which is a random variable that estimates the true expected gradient. Because gradient estimators with lower variance usually lead to faster convergence \cite{bottou-nocedal}, there have been numerous techniques proposed for modifying gradient estimators in such a way that reduces their variance \cite{monte-carlo-gradient}. When the gradient estimator consists of \textit{score functions}, one variance reduction technique that has received considerable attention is that of baselines. 

Our work is motivated primarily by seminal policy gradient algorithms such as A2C \cite{a3c} or PPO \cite{ppo}, where the gradient estimator for the policy consists entirely of score functions. Both of those algorithms also estimate the value function, which is used as a baseline for the policy gradient. And while the success of algorithms like PPO have shown that the value function baseline is clearly a useful tool for variance reduction, ultimately, using the value function as a baseline is a heuristic with seemingly little formal justification. 

This paper is motivated by a simple idea: what if, instead of using the value function as a baseline, we used the optimal baseline---the baseline which results in a gradient estimator with minimum variance? For an arbitrary reinforcement learning problem, we derive the optimal baseline and propose two different formulations for applying optimal baselines to a policy gradient method. We also give sufficient conditions for the value function baseline to be optimal. We test the optimal baselines both on toy problems which can be solved analytically, as well as more practical problems which are solved using deep reinforcement learning. 

Although the examples in this paper are focused on the context of Markov decision processes and policy gradient methods, the results are more generally applicable to any problems which lead to gradient estimators that contain score functions.

\section{Background}
\subsection{Reinforcement Learning}
In this paper we shall mainly be concerned with model-free reinforcement learning (RL) problems. In such problems, we have discrete timesteps $t = 1, \ldots, T$, where each timestep is associated with a tuple $(s_t, a_t, r_t)$ that gives the current timestep's state, action, and reward, respectively. At each timestep, given the current state $s_t \in \mathcal{S}$, the \textit{agent} must choose some admissible action $a_t \in \mathcal{A}$ to undertake. Given the state-action pair $(s_t, a_t)$, the \textit{environment} gives a reward $r_t$ and generates the next state $s_{t+1}$. The goal of the problem is to take actions such that the expected episodic reward 
\begin{align*} 
\mathe \Big[ \sum_{t=1}^T r_t \Big] \stepcounter{equation}\tag{\theequation}\label{obj}
\end{align*}
is maximized. Note that the expectation in \eqref{obj} is with respect to the entire trajectory $(s_1, a_1, r_1), \ldots, (s_T, a_T, r_T)$. Policy gradient methods attempt to solve this RL problem by parameterizing an action distribution for any given state. 

When the action space $\mathcal{A}$ is discrete, we assume a function approximator $\pi_{\theta}$ such that $\pi_{\theta}(a_t \, | \, s_t)$ gives the probability of selecting action $a_t$ while in state $s_t$. The parameters $\theta$ describe the action distribution and must be learnt during training. If the action space is continuous, we have some function approximator (with parameters $\theta$) which outputs the mean and standard deviation of a normal distribution for each dimension of $\mathcal{A}$. For continuous actions, $\pi_{\theta}(a_t \, | \, s_t)$ denotes the probability density of taking action $a_t$ in state $s_t$.

We assume that the parameters $\theta$ will be learnt using gradient-based optimization, and that we have a gradient estimator $\hat g_{\rm sf}$ of the form 
\begin{align*} 
\hat g_{\rm sf} := \sum_{i=1}^n F_i \dfrac{\partial \log \pi_{\theta}(a_i \, | \, s_i)}{\partial \theta} \stepcounter{equation}\tag{\theequation}\label{ghat}
\end{align*}
where the scalar $F_i$ may either be an estimate of the \textit{return}, or an \textit{advantage} (see section \ref{GAE}). Note the separate indices $i=1, \ldots, n$ for indexing $\hat g_{\rm sf}$, because the gradient may consist of a random mini-batch of transitions from different episodes. 

In the simplest case, $F_i$ may be the Monte Carlo return
\begin{align*} 
F_i := \sum_{t=i}^T r_t \stepcounter{equation}\tag{\theequation}\label{reinforce}
\end{align*}
which makes $\hat g_{\rm sf}$ an unbiased gradient estimator. When $F_i$ is defined according to \eqref{reinforce}, we call \eqref{ghat} the reinforce gradient, as it is reminiscent of one of the first policy gradient methods \cite{Williams}. In practice, the reinforce gradient may be infrequently used due to its high variance. Biased gradient estimators, such as temporal difference learning \cite{sutton-88} or the generalized advantage estimator \cite{gae}, are often preferred (section \ref{GAE}). 

\subsection{Score Functions}
The term
\begin{align*} 
\dfrac{\partial \log \pi_{\theta}(a_i \, | \, s_i)}{\partial \theta} \, = \, \dfrac{1}{\pi_{\theta}(a_i \, | \, s_i)}\dfrac{\partial \pi_{\theta}(a_i \, | \, s_i)}{\partial \theta}  \stepcounter{equation}\tag{\theequation}\label{SF}
\end{align*}
is called a score function (or alternatively, a likelihood ratio). The $i$\textsuperscript{th} score function is a vector pointing in the direction of parameter space which most steeply increases the probability of sampling $a_i$. This direction is scaled based on the weight $1 / \pi_{\theta}(a_i \, | \, s_i)$, so that actions which are unlikely to be taken receive additional weight to their gradient updates. Notice that in the actual gradient estimator \eqref{ghat}, the $i$\textsuperscript{th} score function is multiplied by $F_i$, so the sign and magnitude of $F_i$ is what ultimately dictates whether that score function should increase or decrease the probability/probability density, and by how much. 

Gradient terms involving score functions arise when the model parameters affect a probability distribution. In this case, we have the conditional probability distribution $\pi_{\theta}(a_t \, | \, s_t)$ (which obviously depends on $\theta$), so we should expect the gradient to have score function terms. The score function gradient should not be confused with the reparameterization trick gradient \cite{monte-carlo-gradient, stochastic-computation-graphs}. 

In the appendix, we show how to derive the unbiased gradient estimator \eqref{ghat}, \eqref{reinforce} from differentiating the objective \eqref{obj}. We also discuss the inclusion of importance sampling, which is a relevant detail for off-policy RL algorithms.

\subsection{Baselines}
Baselines are a control variate with the form
\begin{align*} 
cv \, := \, \sum_{i=1}^n \beta_{\phi}(\xi_i)\dfrac{\partial \log \pi_{\theta}(a_i \, | \, s_i)}{\partial \theta} \stepcounter{equation}\tag{\theequation}\label{cv}
\end{align*}
where $\beta_{\phi}$ is a baseline function, a scalar function with parameters $\phi$. The baseline function may also accept some inputs which we denote as $\xi_i$. In the context of reinforcement learning, $\xi_i$ may be any subset of $\{ s_1, a_1, r_1, s_2, \ldots, s_i\}$ (appendix \ref{baselines-appendix}). If the baseline accepts no inputs, then we say it is a constant baseline. The most fundamental property of baselines is that they have expectation zero, that is $\mathe \big[ cv \big] = 0$ (appendix \ref{baselines-appendix}). Thus we can define the new gradient estimator 
\begin{align*} 
\hat g := & \ \hat g_{\rm sf} - cv \\
 = & \ \sum_{i=1}^n \Big(F_i - \beta_{\phi}(\xi_i) \Big) \dfrac{\partial \log \pi_{\theta}(a_i \, | \, s_i)}{\partial \theta} . \stepcounter{equation}\tag{\theequation}\label{ghat-actual}
\end{align*}
Note that because $\mathe \big[ \hat g \big] \, = \, \mathe \big[ \hat g_{\rm sf}\big]$, the addition of the baselines does not introduce any bias to the gradient estimator.

Since \eqref{cv} and \eqref{ghat} have similar forms, there is strong potential for correlation between $\hat g_{\rm sf}$ and $cv$. For this reason, if $\phi$ are chosen appropriately, we would expect $\hat g$ to have less variance than $\hat g_{\rm sf}$.
\subsection{Value Function Baselines}
Define $\hat V(s_i)$ to be the state-value function for a state $s_i$
\begin{align*} 
& \hat V(s_i) \, := \,  \mathe \bigg[ \sum_{t=i}^T r_t \ \Big| \ s_i \bigg] . 
\end{align*} 
A value function baseline defines $\xi_i := s_i$ and learns $\phi$ so that $\beta_{\phi}(s_i)$ will be an estimate of $\hat V(s_i)$. Value function baselines are widely used \cite{Williams, sutton-88, a3c, greensmith-2004}, including in contexts outside of reinforcement learning \cite{credit-assignment-techniques, nvil}, and are undoubtedly the most common type of baseline.

Formally, we define a value function baseline as learning $\phi$ by minimizing 
\begin{align*} 
\underset{\phi}{\min} & \quad \sum_{i=1}^n \Big( \mathe \big[ R_i \, | \, s_i \big] - \beta_{\phi}(s_i) \Big)^2 \stepcounter{equation}\tag{\theequation}\label{baseline-obj}
\end{align*}
where $R_i$ is an estimate of the return (e.g. the Monte Carlo return $\sum_{t=i}^T r_t$). 
The objective function \eqref{baseline-obj} gives the gradient
\begin{align*} 
\sum_{i=1}^n -2 \big( R_i - \beta_{\phi}(s_i) \big) \dfrac{\partial \beta_{\phi}(s_i)}{\partial \phi}  \stepcounter{equation}\tag{\theequation}\label{nabla-phi}
\end{align*}
which can be used to update $\phi$ iteratively. Thus, in a typical policy gradient algorithm, we have two function approximators $\pi_{\theta}$ and $\beta_{\phi}$, both of which are updated according to their respective gradients. 
\subsection{Generalized Advantage Estimation} \label{GAE}
The generalized advantage estimator (GAE) \cite{gae} defines 
\begin{align*} 
F_i := \sum_{t=i}^T (\gamma \kappa)^{t-i} r_t + \sum_{t=i}^{T-1} (\gamma \kappa)^{t-i}(1-\kappa)\gamma \hat V(s_{t+1}) - \hat V(s_i) \stepcounter{equation}\tag{\theequation}\label{gae}
\end{align*}
where $\gamma \in (0, 1]$ is a discount factor and $\kappa \in [0, 1]$ is a GAE hyperparameter. Note that Eq.\@ \eqref{gae} is equivalent to the original GAE but we have rearranged terms and considered a finite time MDP.
The first two terms of \eqref{gae},
\begin{align*} 
\sum_{t=i}^T (\gamma \kappa)^{t-i} r_t + \sum_{t=i}^{T-1} (\gamma \kappa)^{t-i}(1-\kappa)\gamma \hat V(s_{t+1}) \stepcounter{equation}\tag{\theequation}\label{return}
\end{align*}
 are an estimate of the (discounted) return $\sum_{t=i}^T \gamma^{t-i}r_t$. The last term of gae, $-\hat V(s_i)$, simply acts as a value function baseline for the return. The generalized advantage estimator is called an \textit{advantage} because it is the difference between a return and a value function baseline. 
 
 In terms of the score function gradient, advantages have a simple interpretation. The value function baseline measures the average return the agent expects following some given state. If the actual return was larger than the expected return, then the advantage is positive. In that case, the score function will increase the probability of its corresponding action. In the opposite case, where the advantage is negative, the probability of taking that action will decrease. Note that in our notation, $F_i$ may either be a return (e.g. reinforce) or an advantage (e.g. GAE).
 
 The two parameters $\gamma, \kappa$ both control the bias-variance trade-off and should ideally be tuned per individual problem. Smaller values of $\gamma$ will reduce the variance of $F_i$, but also bias the agent towards nearsighted actions. The hyperparameter $\kappa$ controls the amount of bootstrapping used in $F_i$, with lower values of $\kappa$ adding more bias but also reducing variance. %If $\kappa=1$, then GAE estimates defined as the discounted Monte Carlo return $\sum_{t=i}^T\gamma^tr_t$. 

\section{The Optimal Baseline}
The gradient estimator $\hat g$ is a function both of the policy parameters $\theta$ and the baseline function parameters $\phi$. We call a baseline optimal if, for some given $\theta$, the parameters $\phi$ minimize the variance of $\hat g$. For the vector-valued $\hat g$, we define the variance as
\begin{align*} 
\text{Var}\big[ \hat g \big] \,:=\, \mathe \big[ \norm{\hat g}^2 \big] - \norm{\mathe \big[ \hat g \big]}^2 \, \stepcounter{equation}\tag{\theequation}\label{var-defn}
\end{align*} 
where $\norm{\cdot}$ denotes the Euclidean norm. Eq.\@ \eqref{var-defn} defines the variance of a vector as the sum of the variance of each of its components. An important observation is that if the baselines have expectation 0, then $\norm{\mathe [ \hat g]}^2$ is constant with respect to $\phi$ (recall that this is true in the usual formulation where $\xi_i := s_i$). In that case, minimizing the variance of $\hat g$ is equivalent to minimizing the second moment $\mathe [ \norm{\hat g}^2]$.

For this reason, we will consider an optimal baseline to be one which minimizes the second moment of $\hat g$. Moreover, works analyzing stochastic gradient descent such as \cite{bottou-nocedal, nemirovski, almost-sure-SGD} all give convergence results which directly depend on the second moment of $\hat g$, so there are strong theoretical justifications for defining optimal baselines as minimizing the second moment of $\hat g$. 

\begin{theorem}\label{thm1}
For any policy $\pi_{\theta}$, the baseline functions which minimize the second moment of Eq.\@ \eqref{ghat-actual} are given by
\begin{align*} 
\beta_{\phi}(\xi_i) = \dfrac{ \mathe \bigg[ \innerp{ \hat g_{\rm sf} , \dfrac{\partial \log \pi_{\theta}(a_i \, | \, s_i)}{\partial \theta}} \ \Big| \  \xi_i \bigg]}{\mathe \bigg[  \innerp{\dfrac{\partial \log \pi_{\theta}(a_i \, | \, s_i)}{\partial \theta}, \dfrac{\partial \log \pi_{\theta}(a_i \, | \, s_i)}{\partial \theta}} \ \Big| \  \xi_i \bigg]} \stepcounter{equation}\tag{\theequation}\label{thm1-eqn}
\end{align*}
where $\innerp{\cdot, \cdot}$ is the inner product, the expectations are with respect to the trajectory $(s_1, a_1, r_1),\ldots,(s_T, a_T, r_T)$, and $\xi_i$ may be any subset of $\{s_1, a_1, r_1, \ldots, s_i\}$. 
\end{theorem}
\textit{Proof.} In order to prove the theorem, we will first show that for any $i, j$ such that $i \neq j$,
\begin{align*} 
\mathe \bigg[ \beta_{\phi}(\xi_i) \innerp{\dfrac{\partial \log \pi_{\theta}(a_i \, | \, s_i)}{\partial \theta}, \dfrac{\partial \log \pi_{\theta}(a_j \, | \, s_j)}{\partial \theta}} \ \Big| \ \xi_j \bigg] \, = \, 0. \stepcounter{equation}\tag{\theequation}\label{thm1-proof1}
\end{align*}
First consider the case that $j > i$; then 
\begin{align*} 
& \mathe_{s_1^T, a_1^T, r_1^T} \bigg[ \beta_{\phi}(\xi_i) \innerp{\dfrac{\partial \log \pi_{\theta}(a_i \, | \, s_i)}{\partial \theta}, \dfrac{\partial \log \pi_{\theta}(a_j \, | \, s_j)}{\partial \theta}} \ \Big| \ \xi_j \bigg] \\
=  \, & \mathe_{s_1^j, a_1^{j-1}, r_1^{j-1}} \bigg[ \mathe_{a_j^T, r_j^T, s_{j+1}^T} \bigg[ \beta_{\phi}(\xi_i) \innerp{\dfrac{\partial \log \pi_{\theta}(a_i \, | \, s_i)}{\partial \theta}, \dfrac{\partial \log \pi_{\theta}(a_j \, | \, s_j)}{\partial \theta}} \ \Big| \ s_1^j, \, a_1^{j-1}, \, r_1^{j-1}, \, \xi_j \, \bigg] \bigg] \\
= \, & \mathe_{s_1^j, a_1^{j-1}, r_1^{j-1}} \bigg[  \beta_{\phi}(\xi_i) \dfrac{\partial \log \pi_{\theta}(a_i \, | \, s_i)}{\partial \theta} \mathe_{a_j^T, r_j^T, s_{j+1}^T} \bigg[\dfrac{\partial \log \pi_{\theta}(a_j \, | \, s_j)}{\partial \theta}^\top \ \Big| \ s_1^j, \, a_1^{j-1}, \, r_1^{j-1} \, \bigg] \bigg]  \\
= \, & \mathe_{s_1^j, a_1^{j-1}, r_1^{j-1}} \bigg[  \beta_{\phi}(\xi_i) \dfrac{\partial \log \pi_{\theta}(a_i \, | \, s_i)}{\partial \theta}  \, ( 0 )  \, \bigg] \ = \ 0\,  \stepcounter{equation}\tag{\theequation}\label{thm1-proof11}
\end{align*}
where the last line uses the baseline property \ref{baselines-appendix}. Note also the notation $s_1^T$ which we use as shorthand for $\{s_1, \ldots, s_T\}$ (and similarly for $a_1^T, r_1^T$). \\
In the other case that $i > j$, then we have essentially the same argument 
\begin{align*} 
& \mathe_{s_1^T, a_1^T, r_1^T} \bigg[ \beta_{\phi}(\xi_i) \innerp{\dfrac{\partial \log \pi_{\theta}(a_i \, | \, s_i)}{\partial \theta}, \dfrac{\partial \log \pi_{\theta}(a_j \, | \, s_j)}{\partial \theta}} \ \Big| \ \xi_j \bigg] \\
= \, & \mathe_{s_1^i, a_1^{i-1}, r_1^{i-1}} \bigg[  \beta_{\phi}(\xi_i) \dfrac{\partial \log \pi_{\theta}(a_j \, | \, s_j)}{\partial \theta} \mathe_{a_i^T, r_i^T, s_{i+1}^T} \bigg[\dfrac{\partial \log \pi_{\theta}(a_i \, | \, s_i)}{\partial \theta}^\top \ \Big| \ s_1^i, \, a_1^{i-1}, \, r_1^{i-1} \, \bigg] \bigg]  \ = \ 0 .
\end{align*}
and so we have established \eqref{thm1-proof1}. \\

\noindent Now setting $\frac{\partial}{\partial \phi} \text{Var}(\hat g) = 0$, we have
\begin{align*} 
& 0 \ = \ \dfrac{\partial }{\partial \phi}\mathe \big[ \norm{\hat g}^2 \big] \\
& = \ -2 \mathe_{s_1^T, a_1^T, r_1^T}\bigg[ \sum_{i=1}^n \innerp{\hat g, \dfrac{\partial \log \pi_{\theta}(a_i \, | \, s_i)}{\partial \theta}} \dfrac{\partial \beta_{\phi}(\xi_i)}{\partial \phi} \bigg] \\
& = \ -2 \mathe_{s_1^T, a_1^T, r_1^T}\bigg[ \sum_{j=1}^n \sum_{i=1}^n \Big( F_i - \beta_{\phi}(\xi_i) \Big) \innerp{\dfrac{\partial \log \pi_{\theta}(a_i \, | \, s_i)}{\partial \theta}, \dfrac{\partial \log \pi_{\theta}(a_j \, | \, s_j)}{\partial \theta}}\dfrac{\partial \beta_{\phi}(\xi_j)}{\partial \phi} \bigg] \\
& = \ -2 \sum_{j=1}^n \mathe_{\xi_j}\bigg[\dfrac{\partial \beta_{\phi}(\xi_j)}{\partial \phi} \mathe_{s_1^T, \, a_1^T, \, r_1^T} \bigg[ \sum_{i=1}^n\Big(F_i - \beta_{\phi}(\xi_i) \Big) \innerp{\dfrac{\partial \log \pi_{\theta}(a_i \, | \, s_i)}{\partial \theta}, \dfrac{\partial \log \pi_{\theta}(a_j \, | \, s_j)}{\partial \theta}} \, \Big| \,  \xi_j  \bigg] \bigg] \\
& = \, -2 \sum_{j=1}^n \mathe_{\xi_j }\bigg[\dfrac{\partial \beta_{\phi}(\xi_j)}{\partial \phi} \mathe_{s_1^T, \, a_1^T, \, r_1^T} \bigg[ \innerp{\hat g_{\rm sf} - \beta_{\phi}(\xi_j)\dfrac{\partial \log \pi_{\theta}(a_j \, | \, s_j)}{\partial \theta}, \dfrac{\partial \log \pi_{\theta}(a_j \, | \, s_j)}{\partial \theta}} \, \Big| \,  \xi_j \bigg] \bigg] \stepcounter{equation}\tag{\theequation}\label{thm1-proof2}
\end{align*}
where the last equality uses Eq.\@ \eqref{thm1-proof1}. 

For any policy $\pi_{\theta}$ and any partial sample path $\xi_j$, there must be some (not necessarily unique) scalar value $\mathcal{B}$ which is the optimal baseline function for the $\frac{\partial \log \pi_{\theta}(a_j \, | \, s_j)}{\partial \theta}$ score function. Realizing that $\frac{\partial \beta_{\phi}(\xi_j)}{\partial \phi}$ merely encodes the direction of steepest ascent for $\beta_{\phi}$, it follows that $\frac{\partial \beta_{\phi}(\xi_j)}{\partial \phi}$ has no bearing whatsoever on the actual value of $\mathcal{B}$. Based on this observation, for \eqref{thm1-proof2} to be equal to zero, the inner expectation of \eqref{thm1-proof2} must be zero: 
\pushQED{\qed}
\begin{align*} 
&  \mathe_{s_1^T, \, a_1^T, \, r_1^T} \bigg[ \innerp{\hat g_{\rm sf} - \beta_{\phi}(\xi_j)\dfrac{\partial \log \pi_{\theta}(a_j \, | \, s_j)}{\partial \theta}, \dfrac{\partial \log \pi_{\theta}(a_j \, | \, s_j)}{\partial \theta}} \, \Big| \,  \xi_j \bigg]  = 0 \\
\Rightarrow \  \ & \beta_{\phi}(\xi_j) \, = \, \dfrac{ \mathe \bigg[ \innerp{ \hat g_{\rm sf} , \dfrac{\partial \log \pi_{\theta}(a_i \, | \, s_i)}{\partial \theta}} \ \Big| \  \xi_i \bigg]}{\mathe \bigg[  \innerp{\dfrac{\partial \log \pi_{\theta}(a_i \, | \, s_i)}{\partial \theta}, \dfrac{\partial \log \pi_{\theta}(a_i \, | \, s_i)}{\partial \theta}} \ \Big| \  \xi_i \bigg]} \, .  \qedhere
\end{align*}
\popQED

\subsection{Discussion of the Optimal Baseline}
The optimal baseline is defined very differently than the value function baseline. Given the widespread use of value function baselines, then, a key questions is ``When can we expect the value function baseline to be optimal or close to optimal?''. 

\begin{theorem} \label{thm2}
Let the mini-batch size be one ($n=1$). For some state $s_i \in \mathcal{S}$, let there exist some constant $K \in \mathbb{R}$ such that 
\begin{align*} 
\innerp{ \dfrac{\partial \log \pi_{\theta}(a_i \, | \, s_i)}{\partial \theta},  \dfrac{\partial \log \pi_{\theta}(a_i \, | \, s_i)}{\partial \theta}} \, = \, K  \stepcounter{equation}\tag{\theequation}\label{cor2-eqn}
\end{align*}
for any $a_i \in \mathcal{A}$. Then, the value function for state $s_i$ is an optimal baseline.
\end{theorem}
\textit{Proof.} Assume \eqref{cor2-eqn} is true and that $n=1$. Without loss of generality, we can assume $\xi_i = \{s_1, a_1, r_1, \ldots, s_i\}$ because putting more information into $\xi_i$ can only increase the variance reduction of the optimal baseline. By definition, 
\begin{align*} 
\hat g_{\rm sf} = R_i \dfrac{\partial \log \pi_{\theta}(a_i \, | \, s_i)}{\partial \theta} \stepcounter{equation}\tag{\theequation}\label{cor2-eqn20}
\end{align*}
where the score function is multiplied by a return denoted as $R_i$, because it is assumed that $\hat g_{\rm sf}$ does not have a baseline. The optimal baseline is 
\begin{align*} 
\beta_{\phi}(\xi_i) \, & = \, \dfrac{ \mathe \bigg[ \innerp{ \hat g_{\rm sf} , \dfrac{\partial \log \pi_{\theta}(a_i \, | \, s_i)}{\partial \theta}} \ \Big| \  \xi_i \bigg]}{\mathe \bigg[  \innerp{\dfrac{\partial \log \pi_{\theta}(a_i \, | \, s_i)}{\partial \theta}, \dfrac{\partial \log \pi_{\theta}(a_i \, | \, s_i)}{\partial \theta}} \ \Big| \  \xi_i \bigg]} \\
& = \, \dfrac{\mathe \bigg[ R_i \innerp{\dfrac{\partial \log \pi_{\theta}(a_i \, | \, s_i)}{\partial \theta}, \dfrac{\partial \log \pi_{\theta}(a_i \, | \, s_i)}{\partial \theta}}  \ \Big| \  \xi_i  \bigg]}{K} \\
& = \, \mathe \Big[ R_i  \, \big| \, \xi_i  \Big] \\
& = \, \mathe \Big[ R_i  \, \big| \, s_i  \Big].  \stepcounter{equation}\tag{\theequation}\label{cor2-eqn2}
\end{align*}
It follows that the value function and optimal baseline have equivalent definitions. \hfill \qed \\

\noindent First, the reason we require $n=1$ is because the optimal baseline is able to take advantage of the correlations between different transitions within a mini-batch. This is also related to the fact that an optimal baseline can use $\xi_i := \{s_1, a_1, r_1, \ldots, s_i\}$, giving extra information on earlier transitions that may be helpful in exploiting those correlations. 

But even when the mini-batch size is one, the optimal baseline will still be distinct from the value function baseline unless \eqref{cor2-eqn} is also satisfied. In practice, we expect that \eqref{cor2-eqn} is hardly ever satisfied: for some policy, there may be certain values of $\theta$ which make \eqref{cor2-eqn} true, but it is difficult to formulate even a toy problem where it holds for an arbitrary $\theta$. One observation, however, is that if \eqref{cor2-eqn} is approximately satisfied, then we would expect the value function baseline to be similar to the optimal baseline. 

%Lastly, we point out that the optimal baseline has a simple geometric interpretation: as the distance such that $\norm{\hat g_{\rm sf} - \beta \dfrac{\partial \log \pi}{\partial \theta}}$ is minimized on average. %%%Don't think this is quite right...but it's definitely suggestive.

\subsection{Function Approximation Optimal Baselines} \label{optimal-baselines}
In special cases, it may be possible to directly calculate \eqref{thm1-eqn} and obtain the optimal baselines in a closed form. In general, we assume that the expectations cannot be directly evaluated, so the optimal baselines should be estimated through an iterative procedure which we will now present. 

We assume that there is a function approximator $\hat \beta$ which outputs two values, $\hat \beta_{\rm top}, \hat \beta_{\rm bot}$, each of which estimates one of the expectations in \eqref{thm1-eqn} (so that $\beta_{\phi} := \hat \beta_{\rm top} / \hat \beta_{\rm bot}$). By considering the objective function 
\begin{align*} 
\underset{\phi}{\min} \ \, & \sum_{i=1}^n\bigg( \mathe \bigg[ \innerp{ \hat g_{\rm sf} , \dfrac{\partial \log \pi_{\theta}(a_i \, | \, s_i)}{\partial \theta}} \ \Big| \  \xi_i \bigg] - \hat \beta_{\rm top}(\xi_i) \bigg)^2 \\
 & + \sum_{i=1}^n \bigg(\mathe \bigg[  \innerp{\dfrac{\partial \log \pi_{\theta}(a_i \, | \, s_i)}{\partial \theta}, \dfrac{\partial \log \pi_{\theta}(a_i \, | \, s_i)}{\partial \theta}} \ \Big| \  \xi_i \bigg] -  \hat \beta_{\rm bot}(\xi_i) \bigg)^2
\end{align*}
we can update $\phi$ using samples of the Monte Carlo gradient
\begin{align*} 
\nabla \phi  = & \sum_{i=1}^n -2\bigg( \innerp{ \hat g_{\rm sf} , \dfrac{\partial \log \pi_{\theta}(a_i \, | \, s_i)}{\partial \theta}} - \hat \beta_{\rm top}(\xi_i) \bigg)\dfrac{\partial \hat \beta_{\rm top}}{\partial \phi} \\
& + \sum_{i=1}^n -2\bigg( \innerp{ \dfrac{\partial \log \pi_{\theta}(a_i \, | \, s_i)}{\partial \theta}, \dfrac{\partial \log \pi_{\theta}(a_i \, | \, s_i)}{\partial \theta}} - \hat \beta_{\rm bot}(\xi_i) \bigg)\dfrac{\partial \hat \beta_{\rm bot}}{\partial \phi}  \stepcounter{equation}\tag{\theequation}\label{optimal-update}
\end{align*}
Note that the optimal baseline requires the computation of all the score functions (i.e. a $n \times \text{dim}(\phi)$ Jacobian matrix). This may be computationally expensive when $n$ is large. Additional implementation details and psuedocode are given in section \ref{ppo}. 
%Algorithm \ref{alg2} also gives psuedocode for a variant of proximal policy optimization \cite{ppo} which uses optimal baselines. 

\subsection{Per-parameter Optimal Baselines} \label{pp-baselines}
For a per-parameter baseline, the baseline function $\beta_{\phi}$ is not a scalar but rather a vector with the same length as $\theta$. When applied to $\hat g_{\rm sf}$, we have
\begin{align*} 
\hat g = \sum_{i=1}^n \Big(F_i \cdot \mathbf{1} - \beta_{\phi}(\xi_i)\Big) \odot \dfrac{\partial \log \pi_{\theta}(a_i \, | \, s_i)}{\partial \theta}
\end{align*}
where $\odot$ denotes element-wise multiplication and $\mathbf{1}$ is a vector of all ones. Thus, each component of the baseline function affects only its corresponding element of $\theta$. The idea of a per-parameter baseline was first proposed in \cite{peters-2008}.

It is simple to extend theorem \ref{thm1} to the optimal per-parameter baseline. Let $\theta_k$ denote the $k$\textsuperscript{th} component of $\theta$, let $\beta_{k}(\xi_i)$ denote the baseline value for $\theta_k$ and define $\big( \hat g_{\rm sf} \big)_k$ as the $k$\textsuperscript{th} component of $\hat g_{\rm sf}$. 
\begin{align*} 
\beta_{k}(\xi_i)  = \dfrac{\mathe \bigg[ \innerp{\big( \hat g_{\rm sf} \big)_k, \dfrac{\partial \log \pi_{\theta}(a_i \, | \, s_i)}{\partial \theta_k}} \ \Big| \ \xi_i \bigg]}{\mathe \bigg[ \innerp{\dfrac{\partial \log \pi_{\theta}(a_i \, | \, s_i)}{\partial \theta_k}, \dfrac{\partial \log \pi_{\theta}(a_i \, | \, s_i)}{\partial \theta_k}} \ \Big| \ \xi_i \bigg]} . \stepcounter{equation}\tag{\theequation}\label{pp-baseline-eqn}
\end{align*}

In our formulation, we let each component of $\theta$ have a single per-parameter baseline which is constant over all states ($\xi_i := \{\}$), so in total there will be $\text{dim}(\theta)$ per-parameter baselines. Each per-parameter baseline needs two parameters (again, one to estimate each of the expectations in \eqref{pp-baseline-eqn}), so we have vectors $\phi_{\rm top}, \, \phi_{\rm bot}$ such that $\beta_{\phi} := \phi_{\rm top} / \phi_{\rm bot}$. The parameters are updated by the Monte Carlo gradients
\begin{align*} 
& \nabla \phi_{\rm top} \, = \, -2 \bigg( \Big(\hat g_{\rm sf}\Big)_k \cdot \dfrac{\partial \log \pi_{\theta}(a_i \, | \, s_i)}{\partial \theta_k} - \phi_{\rm top}\bigg) \\ 
& \nabla \phi_{\rm bot} \, = \, -2 \bigg( \dfrac{\partial \log \pi_{\theta}(a_i \, | \, s_i)}{\partial \theta_k} \cdot \dfrac{\partial \log \pi_{\theta}(a_i \, | \, s_i)}{\partial \theta_k} - \phi_{\rm bot}\bigg)
\end{align*}

\subsection{Other Work on Optimal Baselines}
The recent work \cite{beyond-variance} extended results of \cite{peters-2008}, giving an expression for the optimal state-dependent baseline (appendix \ref{optimal-2021}). However, the theorem \ref{thm1} uses an improved derivation, which eliminates various extraneous terms. In the next section, example \ref{coin-flips} is formulated so that the optimal gradient estimator has zero variance, and we show that only our optimal baseline achieves this. 

\cite{weaver-tao} showed that the optimal constant baseline asymptotically approaches the average reward. That paper also supported the use of value function baselines. \cite{greensmith-2004} gave an algorithm for estimating the optimal constant baseline, and also compared the optimal constant baseline to the value function baseline. It was found that the value function baseline gave better variance reduction. Our theorem \ref{thm2} has shown that the value function baseline should not be regarded as optimal except in very special situations.

\section{Toy Examples}
\subsection{Coin Flips Example} \label{coin-flips}
Consider a game where a gambler flips a coin twice. They are paid \$1 for flipping two tails; \$2 for flipping two heads; or \$4 for flipping one heads and one tails. The gambler provides their own (not necessarily fair) coin, so the question is, with what probability should the coin show heads to maximize the gambler's return? It is simple to show that the best probability is $3/5$, but suppose we wanted to find this value using a policy gradient method. Our parameters are the two logits, $\theta_1, \, \theta_2$, which define the probability of tails/heads through a softmax distribution,
\begin{align*} 
\mathbb{P}(\text{tails}) = \dfrac{e^{\theta_1}}{e^{\theta_1} + e^{\theta_2}} \qquad \mathbb{P}(\text{heads}) = \dfrac{e^{\theta_2}}{e^{\theta_1} + e^{\theta_2}} \, .
\end{align*}
We let each gradient estimator consist of an entire episode (two flips) and let $F_i$ be the Monte Carlo return.

\begin{figure}[H] 
\centering 
\includegraphics[ width=\textwidth]{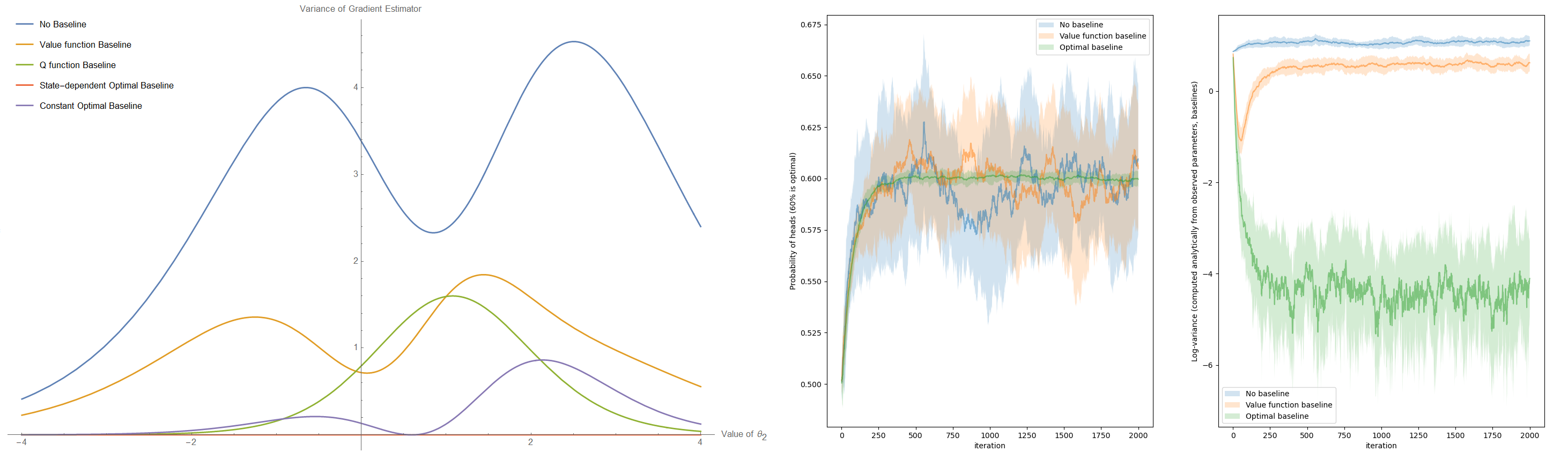}  
\caption{Left panel: analytic plots of the variance of the gradient estimator as a function of $\theta_2$ ($\theta_1 = 1$), for different baseline types. Middle and right panels: numerical plots of the probability of heads, and variance of the gradient estimator during training.} \label{coinflipfigs} 
\end{figure}

Because of its simplicity, the value function, action-value function (Q function), optimal baselines, and variance of the gradient estimator can all be computed analytically. The left panel of figure \ref{coinflipfigs} compares the value function baseline and Q function baseline \eqref{optimal-app-1} with a constant optimal baseline and a state-dependent optimal baseline. Note that the constant optimal baseline has a single baseline value, with the others having three distinct baseline values. 
This example is specially formulated so that it is possible for the gradient estimator to have zero variance. This means that no matter which sample path is observed (heads-heads, heads-tails, tails-heads, or tails-tails), the gradient estimator with optimal baseline is the same as the true expected gradient. The sample paths heads-tails and tails-heads are identical in this case, so to achieve zero variance, the gradient estimator must give the same result, in both of its two components, for each of the 3 unique sample paths. Thus, achieving zero variance requires satisfying a system of 6 equations. In general, we would then need 6 variables: this is precisely how many unique values a state-dependent per-parameter baseline can take. In this case, the optimal per-parameter baseline is actually the same as the optimal baseline (because the per-parameter baseline has the same value in both components), and this is why the optimal baseline can achieve zero variance. In normal problems, achieving zero variance is impossible because of the constraints on $\xi_i$. 

Note that by scaling the possible rewards by a constant factor, the variance of the non-optimal baselines can be made arbitrarily high, but the optimal state-dependent baseline will always have zero variance.

The middle and right panels in figure \ref{coinflipfigs} show the results of applying SGD with a learning rate of $0.01$ starting from an initial guess of $\theta_1, \theta_2 = 1, 1$. In each SGD iteration, we sample an episode and then perform the policy gradient update \eqref{ghat-actual} and a baseline update. The experiment is repeated 20 times and the the mean $\pm$ standard deviation is plotted. The extra variance reduction from the optimal baseline had a clear benefit of keeping the iterates of SGD closer to the optimum value.

\subsection{Multi-Arm Bandit Problem} \label{mab}
Consider a bandit problem with three arms, where each of the arms deterministically gives a constant reward of 0, 0.7, and 1, respectively. Then, the optimal policy is to simply always pull the arm giving a reward of 1. Let the parameters $\theta_1, \, \theta_2, \,  \theta_3$ give the logits for a softmax distribution over the three arms. As the problem has only a single action per episode, we can consider only constant baselines. Thus we may have a single baseline value, or three baseline values when using a per-parameter baseline. Having a zero variance gradient estimator is then impossible as it would require nine baseline values. 

\begin{figure}[H] 
\centering 
\includegraphics[ width=\textwidth]{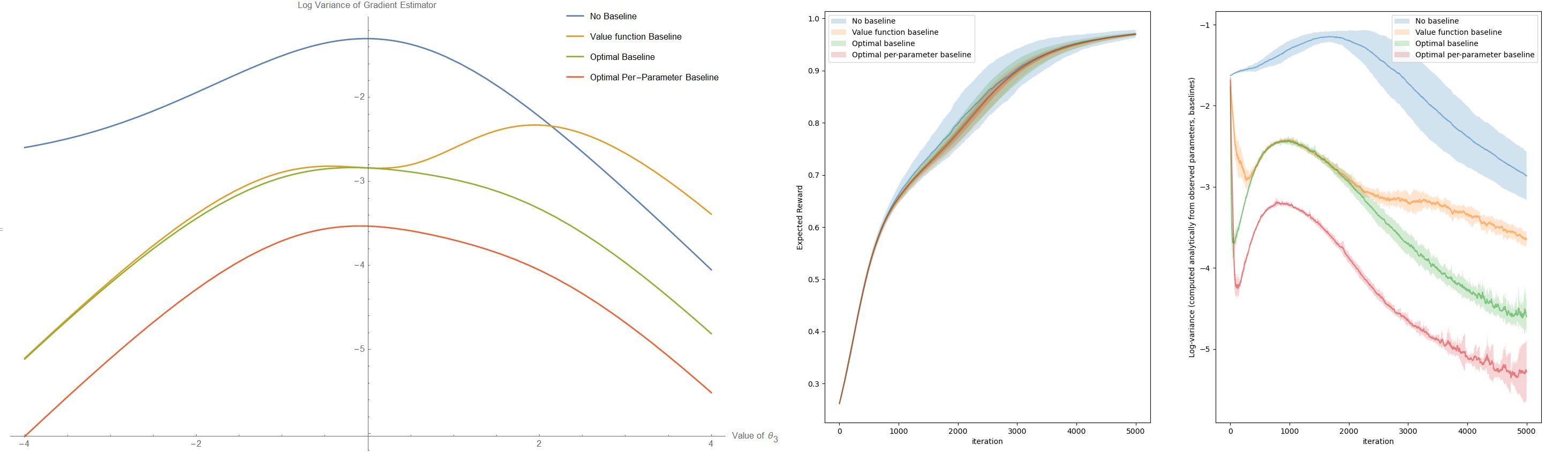} 
\caption{Left panel: analytic plots of variance of the gradient estimator as a function of $\theta_3$ ($\theta_1, \theta_2 = 0$). Middle and right panel: numerical plots of expected reward and variance of the gradient during training.} \label{mabfig}
\end{figure}

Again the different baseline values and their corresponding variance are computed analytically, and plotted in the left panel of figure \ref{mabfig} for varying parameter values. The middle and right panels show the numerical solutions when starting from an initial value of $\theta_1, \theta_2, \theta_3 = 3,2,1$ (favoring the 0 reward and 0.7 reward arms) with a learning rate of $0.01$. Again we find that the optimal baseline results in a lower variance compared to the value function baseline, and that the per-parameter baseline further improves on the optimal baseline by roughly an order of magnitude. However, the lower variance did not give any benefits in terms of the convergence of SGD, as all baseline types converged to the optimum at the same rate. 

\subsection{A Simple MDP} \label{mdp}
\begin{figure}[H] 
\centering 
\includegraphics[ width=.8\textwidth]{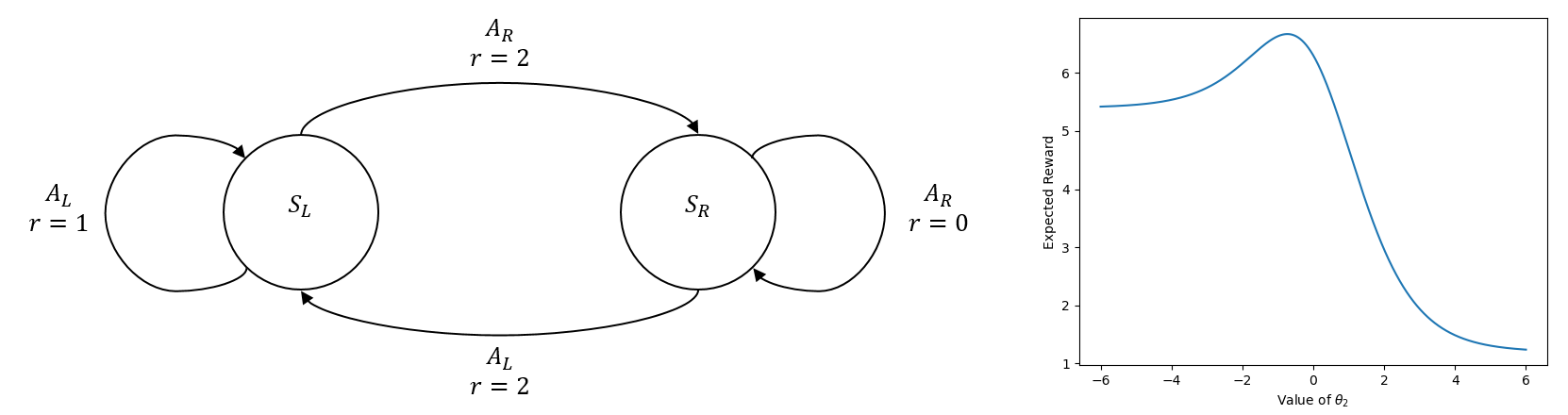} 
\caption{Left: diagram of the two states and the four possible transitions and their associated rewards. Right: expected reward of the MDP, where the initial state distribution is $[ 60\% S_L, 40\% S_R]$ and the discount factor is $\gamma = 0.8$ (20\% chance of episode termination after any action). Plot shows varying $\theta_2$, with constant $\theta_1 = 0$.} \label{mdp-obj}
\end{figure}

We consider the Markov decision process (MDP) proposed by \cite{russo-policy-gradient}, shown in figure \ref{mdp-obj}. This is a simple MDP with two states (denoted as $S_L$ and $S_R$). There are always two actions possible (denoted as $A_L$ and $A_R$) which are parameterized with a softmax distribution with associated logits $\theta_1$ and $\theta_2$ respectively. The policy chooses between the actions without regard to the current state.
  
\begin{figure}[H] 
\centering 
\includegraphics[ width=\textwidth]{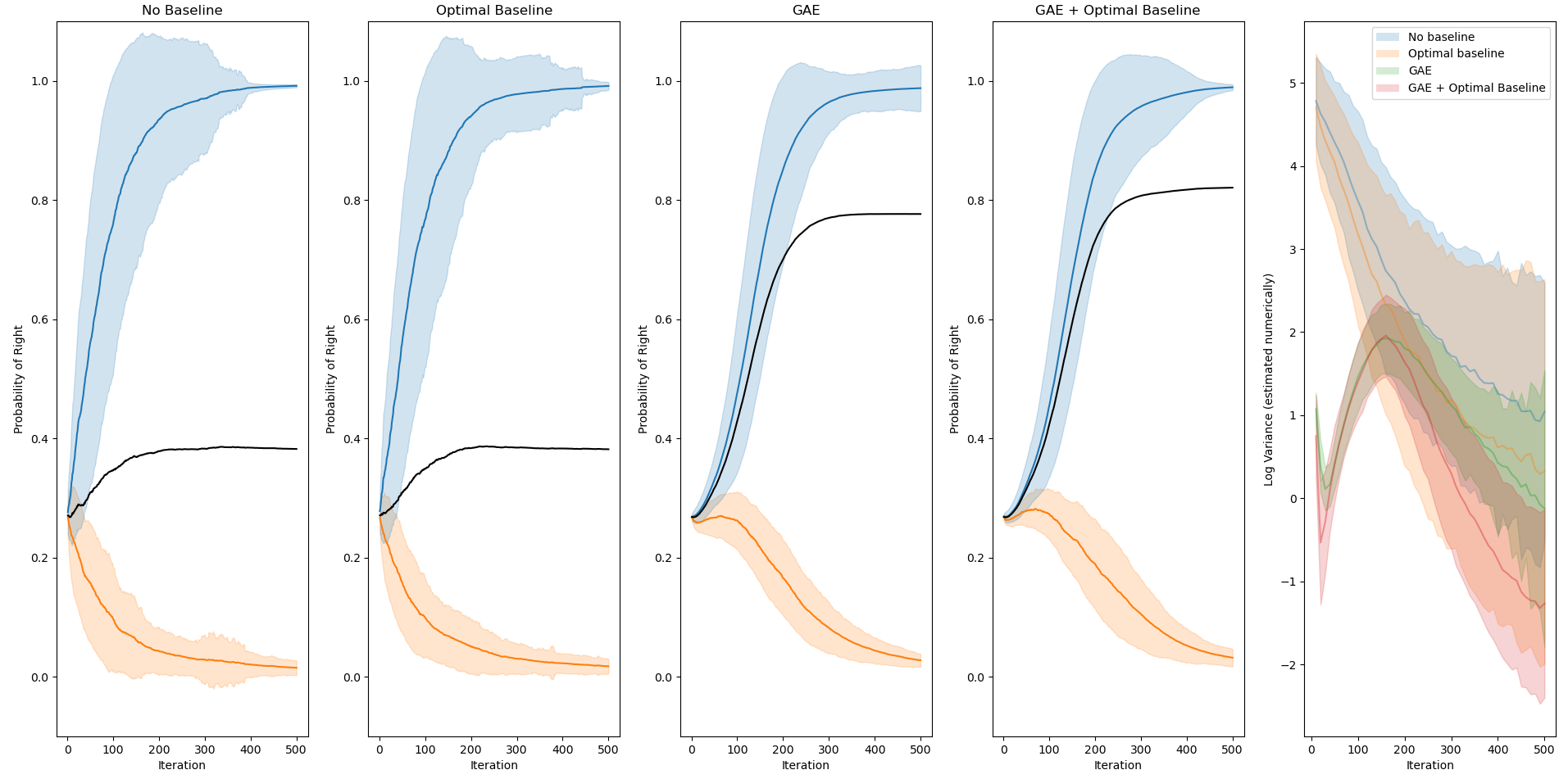} 
\caption{Left four panels: Over 500 replications, trajectories are grouped according to whether they converge to the suboptimal (orange) or optimal (blue) policy; the training curves for each group of trajectories are shown. The black line shows the average over all trajectories. Right panel: average observed variance during training for the different baselines.} \label{mdp-fig}
\end{figure}

Here we aim to \textit{minimize} the expected reward, so the optimal policy is to simply always move right (always choose $A_R$). The interesting thing about this example is that the expected reward has two minima corresponding to always choosing $A_L$, or always choosing $A_R$ (right panel of figure \ref{mdp-obj}). If one follows the expected gradient, starting with any $\theta_2 < \theta_1 - \log(27/13) \approx \theta_1 - 0.75$ will result in convergence to the suboptimal policy. Starting from $[\theta_1, \theta_2] = [0, -1]$, we test the effect of baselines on whether we will converge to the suboptimal policy when following the gradient estimator. The results are shown in figure \ref{mdp-fig}.

Notably, GAE ($\kappa = 0.2, \gamma= 0.9$) has a much lower variance immediately after initialization due to its use of bootstrapping. Following the expected gradient immediately after initialization would result in convergence to the suboptimal policy, so the fact that GAE is much better at avoiding the suboptimal policy suggests that it is actually biased towards the optimal policy. When combining GAE with an additional optimal baseline, we observed further variance reduction and a small additional benefit in terms of avoiding the suboptimal policy ($p = .08$ with a chi-squared test). We also tested using an additional state-dependent baseline with GAE, but no extra variance reduction was observed in that case.

\section{Application to Proximal Policy Optimization} \label{ppo}
We now consider solving the Bipedal Walker v3 and Lunar Lander v2 environments, both of which are included in OpenAI gym \cite{gym}. These are somewhat challenging tasks and allow us to test the optimal baseline in both a continuous (bipedal walker) and a discrete (lunar lander) action space. First, we use a standard implementation of proximal policy optimization (PPO) \cite{ppo} with tuned hyperparameters to solve either task. We then compare vanilla PPO with three different variants of PPO:
\begin{itemize}
\item \textbf{PPO + optimal baseline:} We still have a neural network which learns the value function, but we also have a separate neural network with two outputs which learns the optimal baseline using the gradient update \eqref{optimal-update}. The optimal baseline takes as input the most recent state ($\xi_i := s_i$), and is updated once per mini-batch, just like the policy and value function. We still use the GAE, so the value function is used both for bootstrapping as well as a baseline. Thus, we have two baselines, and really the `optimal baseline' learns the difference between the value function and the optimal baseline for the GAE return. The value function and optimal baseline use the same network architecture, with the exception of the output layer.
\item \textbf{PPO + per-parameter baseline:} Instead of a state-dependent optimal baseline, we use a constant per-parameter optimal baseline as described in section \ref{pp-baselines}. This does not require a function approximator, but adds $2\, \text{dim}(\theta)$ parameters where $\text{dim}(\theta)$ is the number of policy parameters.
\item \textbf{PPO + extra baseline:} This variant serves to test for any additional benefits which may occur due to the extra parameters introduced by the optimal/per-parameter baselines. We have an extra state-dependent, neural network baseline which is applied to the GAE. This extra baseline attempts to learn the advantage, not the return. If the true value function is used in GAE, the extra baseline should theoretically be zero, but if the value function is poorly fit, then the extra baseline may provide significant variance reduction \cite{mirage-action}.
\end{itemize}
Algorithm \ref{vanilla-ppo} gives psuedocode for vanilla PPO, which can be compared to algorithm \ref{optimal-ppo} which implements PPO with an optimal baseline. The code is available at \url{https://github.com/ronan-keane/PPO-tf2}. 

\begin{figure}[H] 
\centering 
\includegraphics[ width=.9\textwidth]{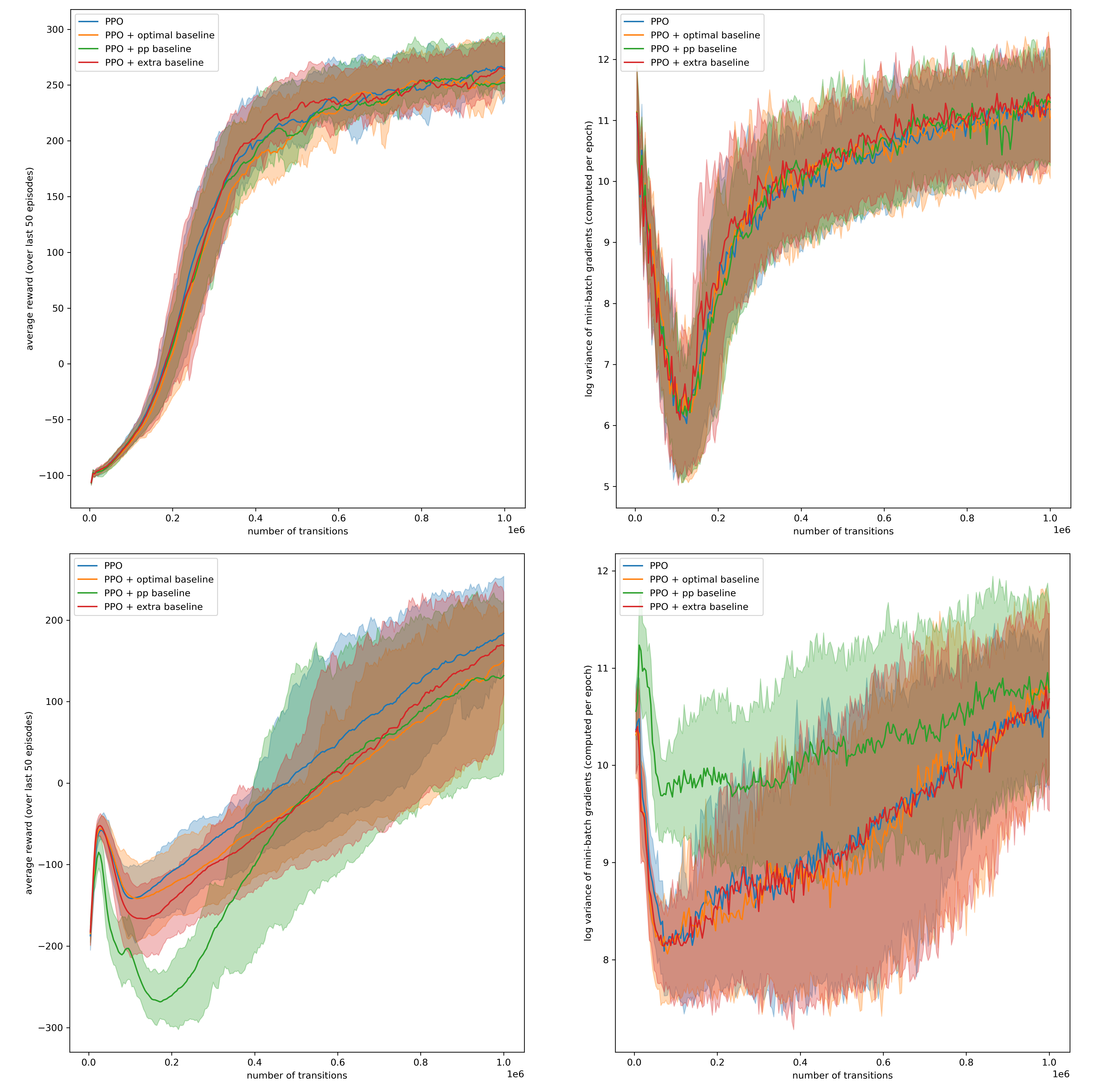} 
\caption{Top: Average reward and variance on Bipedal Walker v3. Bottom: Average reward and variance on Lunar Lander v2. Shaded regions show the 20\textsuperscript{th} and 80\textsuperscript{th} percentiles over 50 replications.}  \label{ppo-results}
\end{figure}

Each of the four algorithms is replicated 50 times on both environments, and the results are shown in figure \ref{ppo-results}. On bipedal walker (top), we see no meaningful difference between the different algorithms, with all algorithms showing similar training curves and similar variance of the policy gradient. It is not surprising that PPO and PPO + extra baseline are extremely similar, but it was unexpected that neither the optimal baseline nor per-parameter baseline provided any tangible improvement. In lunar lander (bottom), the results are even more unexpected. Even though PPO, PPO + optimal baseline and PPO + extra baseline all had extremely similar policy gradient variance during training, each algorithm has a distinct training curve with vanilla PPO learning slightly faster than the other two. For an unknown reason, PPO + per-parameter baseline had a considerably higher variance, but despite this, it still learns at a rate which is comparable to the other algorithms. These results lead us to a similar conclusion as \cite{beyond-variance}, as it is apparent that the connection between the gradient estimator's variance and the convergence rate of the algorithm is not as simple as what theoretical results, such as those of \cite{bottou-nocedal} or \cite{almost-sure-SGD}, may suggest. 

\begin{algorithm}
\caption{Proximal Policy Optimization} \label{vanilla-ppo}
\begin{algorithmic}
\State \textbf{Requires:} Policy $\pi$ with parameters $\theta$, Value function $\hat V$ with parameters $\psi$, learning rates $\alpha_{\theta}, \alpha_{\psi}$, discount factor $\gamma$, GAE parameter $\kappa$, PPO clipping parameter $\epsilon$
\State \vspace{-2mm}
\For{iteration in $niterations$}
\For{$i$ in range($nsteps$) }
\State Sample $a_i$ according to policy; Record $(a_i, r_i, s_{i+1})$ transition
\State Keep $\pi_{\theta_{\rm old}}(a_i \, | \, s_i)$ in memory
\EndFor
\For{epoch in $nepochs$ }
\State For all transitions, compute $F_i$ according to GAE \eqref{gae} 
\State For all states $s_i$, compute $\text{return}(s_i) = F_i + \hat V(s_i)$
\For{each mini-batch in epoch}
\State $IS_i = \frac{\pi_{\theta}(a_i \, | \, s_i)}{\pi_{\theta_{\rm old}}(a_i \, | \, s_i)}$  \Comment{Importance Sampling Ratio}
\State $CLIP_i = 1 - \mathbbm{1}\big((F_i > 0, \, IS_i > 1 + \epsilon) \text{ or } (F_i < 0, \, IS_i < 1 - \epsilon) \big)$ \Comment{$\mathbbm{1}$ is indicator function}
\State $\hat g = \sum_{i=1}^n F_i \cdot IS_i \cdot CLIP_i \frac{\partial \log \pi_{\theta}(a_i \, | \, s_i)}{\partial \theta}$
\State $\theta \gets \theta + \alpha_{\theta} \hat g$
\State $\nabla \psi = -2\sum_{i=1}^n\big(\text{return}(s_i) - \hat V(s_i) \big) \frac{\partial \hat V(s_i)}{\partial \psi}$
\State $\psi \gets \psi - \alpha_{\psi} \nabla \psi$
\EndFor
\EndFor
\EndFor
\end{algorithmic}
\end{algorithm}

\begin{algorithm}
\caption{Proximal Policy Optimization with Optimal Baseline} \label{optimal-ppo}
\begin{algorithmic}
\State \textbf{Requires:} Policy $\pi$ with parameters $\theta$, Value function $\hat V$ with parameters $\psi$, Optimal baseline $\hat \beta$ with parameters $\phi$, learning rates $\alpha_{\theta}, \alpha_{\psi}, \alpha_{\phi}$, discount factor $\gamma$, GAE parameter $\kappa$, PPO clipping parameter $\epsilon$
\State \vspace{-2mm}
\For{ iteration in $niterations$}
\For{$i$ in range($nsteps$) }
\State Sample $a_i$ according to policy; Record $(a_i, r_i, s_{i+1})$ transition
\State Keep $\pi_{\theta_{\rm old}}(a_i \, | \, s_i)$ in memory
\EndFor
\For{epoch in $nepochs$ }
\State For all transitions, compute $F_i$ according to GAE \eqref{gae} 
\State For all states $s_i$, compute $\text{return}(s_i) = F_i + \hat V(s_i)$
\State For all states $s_i$, compute $\beta_{\phi}(s_i) = \hat \beta_{\rm top}(s_i) / \hat \beta_{\rm bot}(s_i)$
\For{each mini-batch in epoch}
\State Compute $\frac{\partial \log \pi_{\theta}(a_i \, | \, s_i)}{\partial \theta}$ for each $i = 1, \ldots, n$ in mini-batch \Comment{compute Jacobian}
\State $IS_i = \frac{\pi_{\theta}(a_i \, | \, s_i)}{\pi_{\theta_{\rm old}}(a_i \, | \, s_i)}$ 
\State $CLIP_i = 1 - \mathbbm{1}\big((F_i > 0, \, IS_i > 1 + \epsilon) \text{ or } (F_i < 0, \, IS_i < 1 - \epsilon) \big)$ 
\State $\hat g = \sum_{i=1}^n \big(F_i - \beta_{\phi}(s_i)\big) \cdot IS_i \cdot CLIP_i \frac{\partial \log \pi_{\theta}(a_i \, | \, s_i)}{\partial \theta}$ 
\State $\theta \gets \theta + \alpha_{\theta} \hat g$ 
\State $\nabla \psi = -2\sum_{i=1}^n\big(\text{return}(s_i) - \hat V(s_i) \big) \frac{\partial \hat V(s_i)}{\partial \psi}$
\State $\psi \gets \psi - \alpha_{\psi} \nabla \psi$
\State $\hat g_{\rm sf} = \sum_{i=1}^n  F_i \cdot IS_i \cdot CLIP_i \frac{\partial \log \pi_{\theta}(a_i \, | \, s_i)}{\partial \theta}$ 
\State $\nabla \phi  = \sum_{i=1}^n -2\Big( \innerp{ \hat g_{\rm sf} , \, IS_i \frac{\partial \log \pi_{\theta}(a_i \, | \, s_i)}{\partial \theta}} - \hat \beta_{\rm top}(s_i) \Big)\frac{\partial \hat \beta_{\rm top}}{\partial \phi}$
\State \hphantom{$\nabla \phi  =$} $+\sum_{i=1}^n -2\Big( \innerp{ IS_i \frac{\partial \log \pi_{\theta}(a_i \, | \, s_i)}{\partial \theta}, \, IS_i \frac{\partial \log \pi_{\theta}(a_i \, | \, s_i)}{\partial \theta}} - \hat \beta_{\rm bot}(s_i) \Big)\frac{\partial \hat \beta_{\rm bot}}{\partial \phi}$
\State $\phi \gets \phi - \alpha_{\phi} \nabla \phi$
\EndFor
\EndFor
\EndFor
\end{algorithmic}
\end{algorithm}

To further investigate optimal baselines, we consider an experiment where we fit the baselines while the policy parameters are held fixed. This removes the effect that the constantly changing policy (which is updated every mini-batch) may have on estimating the baseline values. The algorithms used are exactly the same as the previous experiments, with the exception that the policy update is removed from each mini-batch.

On each of bipedal walker and lunar lander, we saved 12 different policies with varying average rewards, spanning from near the beginning, to the end of training. Then on each policy, we measure the variance of the gradient estimator when using GAE ($\gamma = 0.992, \kappa = 0.5$), GAE with the optimal baseline, and GAE with a per-parameter baseline. For comparison purposes, we also test using reinforce (GAE with $\kappa = 1$ and no value function baseline), reinforce with a value function baseline, and reinforce with an optimal baseline. Each baseline is trained on 100,000 transitions, and then the variance is computed empirically based on another 100,000 transitions. The experiment is replicated 10 times.

\begin{figure}[H] 
\centering 
\includegraphics[ width=.9\textwidth]{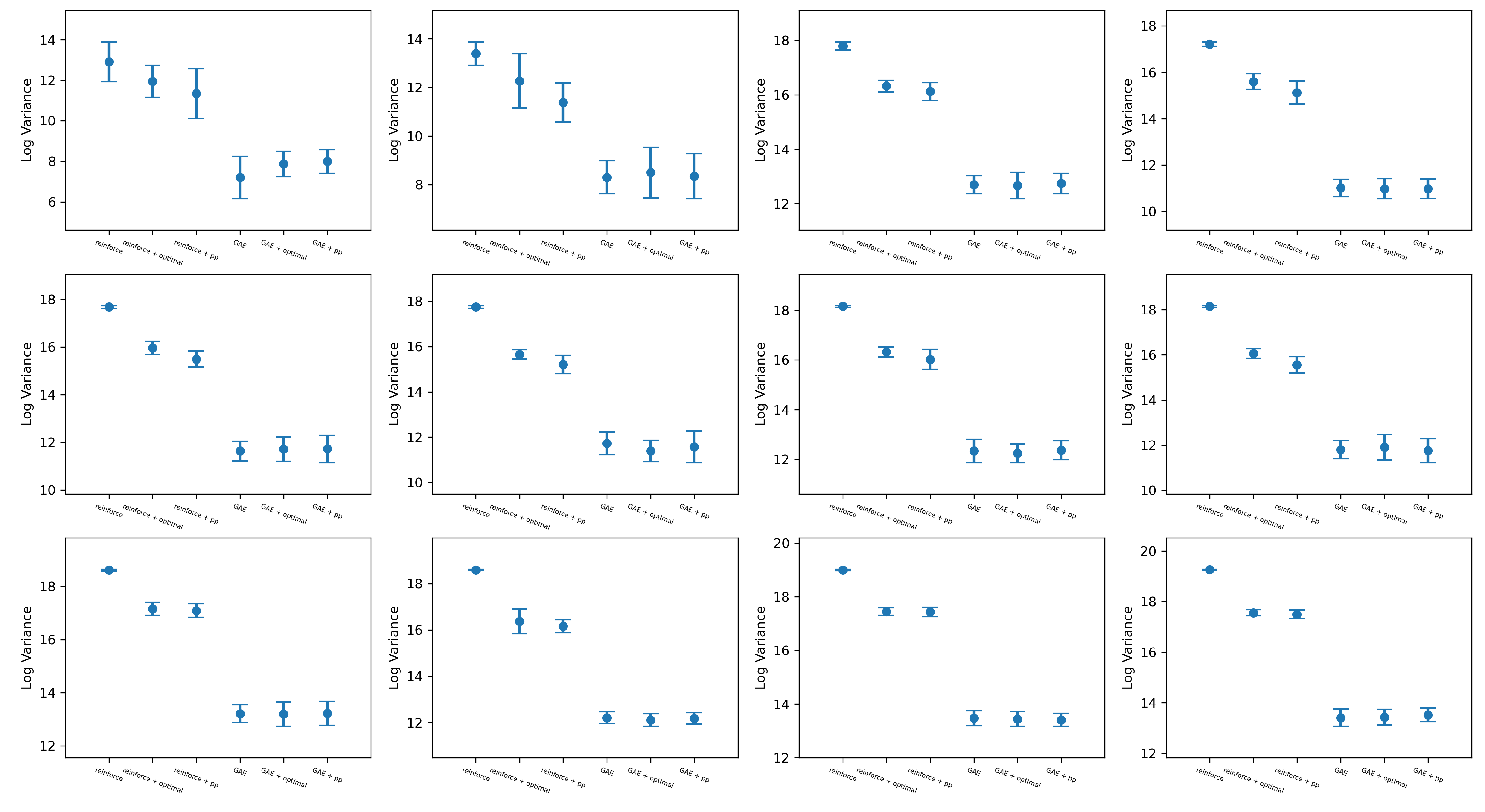} 
\caption{Each plot corresponds to a fixed policy for Bipedal Walker v3. From left to right, the data points show the variance of the policy gradient when using reinforce, reinforce with an optimal baseline, reinforce with a value function baseline, GAE, GAE + optimal baseline, and GAE with per-parameter baseline. Whiskers show the standard deviation of the variance, computed over 10 replications.} \label{walker-fixed}
\end{figure}

\begin{figure}[H] 
\centering 
\includegraphics[ width=.9\textwidth]{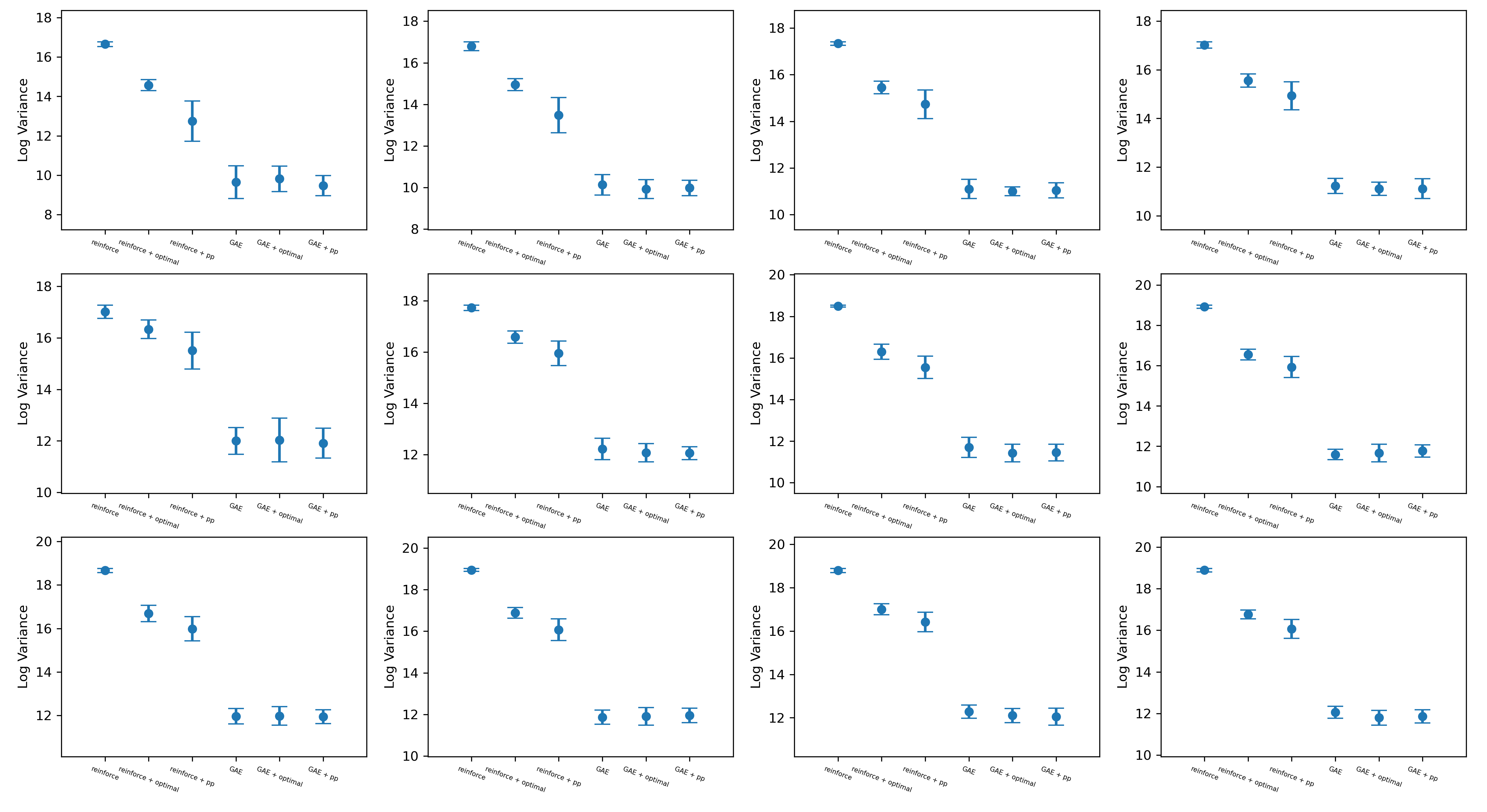} 
\caption{Fixed policy experiments for Lunar Lander v2.} \label{lunar-fixed}
\end{figure}

The results are shown in figures \ref{walker-fixed} and \ref{lunar-fixed}. Before running the experiment, we theorized that reinforce with an optimal baseline should give the same, or slightly lower variance compared to reinforce with a value function baseline. The fact that reinforce with the value function baseline actually gives significantly lower variance suggests that learning the optimal baseline is challenging compared to learning the value function, and that there is a danger of the optimal baseline being underfit in practical problems. These experiments have also shown that the variance reduction from bootstrapping is actually far greater than the variance reduction from baselines. Reinforce by itself only has variance reduction from discounting, so the difference between reinforce and reinforce + value function shows the variance reduction from the value function baseline. Then the difference between reinforce + value function and GAE is the variance reduction from bootstrapping. Although this shows the critical importance of bootstrapping, we must also not forget that bootstrapping adds bias to the gradient, whereas the  baselines are unbiased. 

As for the improvement from the addition of the optimal baselines, on bipedal walker, a paired t-test shows no significant improvement from the optimal baselines (p=0.3021) or per-parameter baselines (p=0.4994). On lunar lander, a paired t-test shows a modest improvement for both the optimal baselines (p=0.04319) and per-parameter baselines (p=0.01066). 

\section{Conclusion}
Using the value function as a baseline has been both a popular and successful choice of baseline, and is often regarded as a crucial part of classic policy gradient algorithms such as A2C/A3C, TRPO, or PPO. Despite this, the value function baseline is fundamentally a heuristic. Previous works arguing for value function baselines have focused on their benefits over constant optimal baselines. In this paper, we have focused on state-dependent optimal baselines, giving an explicit formula for the optimal baselines as well as sufficient conditions for the value function baseline to be optimal. 

We showed that the gap, in terms of variance reduction, between a value function baseline and the optimal baseline may be arbitrarily large. However, in practical problems, the difficulty of learning the optimal baseline (when compared with learning the value function), combined with the fact that the optimal baseline requires Jacobian computations, makes optimal baselines rather impractical. The improvement from the optimal baseline may simply be too small to make a meaningful difference. Additionally, because the variance reduction from bootstrapping is actually larger than the variance reduction from baselines, any practical algorithm would still require a function approximator for the value function. 

\section*{Acknowledgments}
The contents of this report reflect the views of the authors, who are responsible for the facts and the accuracy of the information presented herein. This document is disseminated in the interest of information exchange. The report is funded, partially or entirely, by a grant from the U.S. Department of Transportation’s University Transportation Centers Program. However, the U.S. Government assumes no liability for the contents or use thereof.

\appendix
\section*{Appendix}
\addcontentsline{toc}{section}{Appendices}
\renewcommand{\thesubsection}{\Alph{subsection}}
\subsection{Derivation of Eq.\@ \eqref{ghat}}
Let us assume that the state space, action space and rewards are all continuous. To make the notation as simple as possible, let us assume that $p_{\rm env}(r_t, s_{t+1} \, | \, s_t, a_t)$ gives the joint probability density of receiving reward $r_t$ and transition $s_{t+1}$, given the current state-action pair $(s_t, a_t)$. Also for simplicity, assume that the initial state is deterministic.

Starting with the definition of the expected objective:
\begin{align*} 
& \mathe\Big[ \sum_{t=1}^T r_t \Big] := \mathe_{(s_1, a_1, r_1), \ldots, (s_t, a_t, r_t)} \Big[ \sum_{t=1}^T r_t \Big] \\
:= & \underset{a_1}{\int}\pi_{\theta}(a_1 \, | \, s_1) \underset{r_1, s_2}{\int}p_{\rm env}(r_1, s_2 \, | \, s_1, a_1) \ldots  \underset{a_T}{\int}\pi_{\theta}(a_T \, | \, s_T) \underset{r_T, s_{T+1}}{\int}p_{\rm env}(r_T, s_{T+1} \, | \, s_T, a_T)\Big[ \sum_{t=1}^T r_t \Big] ds_{T+1}dr_T da_T \ldots ds_2dr_1 da_1\\
= & \sum_{t=1}^T \underset{a_1}{\int}\pi_{\theta}(a_1 \, | \, s_1) \underset{r_1, s_2}{\int}p_{\rm env}(r_1, s_2 \, | \, s_1, a_1) \ldots  \underset{a_t}{\int}\pi_{\theta}(a_t \, | \, s_t) \underset{r_t, s_{t+1}}{\int}p_{\rm env}(r_t, s_{t+1} \, | \, s_t, a_t)\big[r_t \big] ds_{t+1}dr_t \ldots da_1 \stepcounter{equation}\tag{\theequation}\label{app-eqn1}
\end{align*}
Now take the derivative of \eqref{app-eqn1} with respect to $\theta$, keeping in mind that $p_{\rm env}(\cdot)$ does not depend on $\theta$.
\begin{align*} 
= &  \sum_{t=1}^T\sum_{j=1}^t \underset{a_1}{\int}\pi_{\theta}(a_1 \, | \, s_1) \ldots \underset{r_{j-1}, s_j}{\int}p_{\rm env}(r_{j-1}, s_j \, | \, s_{j-1}, a_{j-1}) \underset{a_j}{\int} \dfrac{\partial \pi_{\theta}(a_j \, | \, s_j)}{\partial \theta}\ldots \underset{r_{t}, s_{t+1}}{\int}p_{\rm env}(r_{t}, s_{t+1} \, | \, s_{t}, a_{t})\big[ r_t\big] ds_{t+1}dr_t \ldots da_1 \\
= & \sum_{t=1}^T\sum_{j=1}^t \underset{a_1}{\int}\pi_{\theta}(a_1 \, | \, s_1) \ldots \underset{r_{j-1}, s_j}{\int}p_{\rm env}(r_{j-1}, s_j \, | \, s_{j-1}, a_{j-1}) \underset{a_j}{\int} \dfrac{\partial \log \pi_{\theta}(a_j \, | \, s_j)}{\partial \theta}\pi_{\theta}(a_j \, | \, s_j)\ldots \\
& \qquad \ldots \underset{r_{t}, s_{t+1}}{\int}p_{\rm env}(r_{t}, s_{t+1} \, | \, s_{t}, a_{t})\big[ r_t\big] ds_{t+1}dr_t \ldots da_1 \stepcounter{equation}\tag{\theequation}\label{app-eqn2}
\end{align*}
Now to finish the derivation, notice that \eqref{app-eqn2} can be written as an expectation:
\begin{align*} 
= & \sum_{t=1}^T \sum_{j=1}^t \mathe \Big[ r_t \dfrac{\partial \log \pi_{\theta}(a_j \, | \, s_j)}{\partial \theta}\Big] \ \ = \ \ \sum_{j=1}^T\sum_{t=j}^T \mathe \Big[ r_t \dfrac{\partial \log \pi_{\theta}(a_j \, | \, s_j)}{\partial \theta}\Big]  \\
= & \sum_{j=1}^T \mathe \Big[ \sum_{t=j}^T r_t \dfrac{\partial \log \pi_{\theta}(a_j \, | \, s_j)}{\partial \theta}\Big]  \stepcounter{equation}\tag{\theequation}\label{app-eqn3}
\end{align*}
Now comparing the gradient estimator \eqref{ghat} with \eqref{app-eqn3}, we see that they are equal if $T=n$ and $F_i = \sum_{t=i}^T r_t$. Thus, \eqref{ghat} is merely a more general form of the unbiased gradient estimator \eqref{app-eqn3}.

\subsection{Importance Sampling}
In the on-policy setting, given any given state $s_t$, we choose action $a_t$ with probability density $\pi_{\theta}(a_t \, | \, s_t)$. In the off-policy setting, we suppose action $a_t$ was actually chosen according to some different sampling distribution. For example, in proximal policy optimization (PPO), the action $a_t$ has the probability density $\pi_{\theta_{\rm old}}(a_t \, | \, s_t)$, for some previous parameter values $\theta_{\rm old}$. In this case, rather than use the normal score function \eqref{SF}, the score function is multiplied by the importance sampling ratio 
\begin{align*} 
\dfrac{\pi_{\theta}(a_t \, | \, s_t)}{\pi_{\theta_{\rm old}}(a_t \, | \, s_t)}
\end{align*}
which gives the modified score function 
\begin{align*} 
\dfrac{\pi_{\theta}(a_t \, | \, s_t)}{\pi_{\theta_{\rm old}}(a_t \, | \, s_t)} \dfrac{\partial \log \pi_{\theta}(a_t \, | \, s_t)}{\partial \theta}. \stepcounter{equation}\tag{\theequation}\label{IS-SF} 
\end{align*}
Note that it is possible to simplify \eqref{IS-SF} as
\begin{align*} 
\dfrac{1}{\pi_{\theta_{\rm old}}(a_t \, | \, s_t)} \dfrac{\partial \pi_{\theta}(a_t \, | \, s_t)}{\partial \theta} .
\end{align*}
In other words, when using importance sampling, the score function still points in exactly the same direction, but it is scaled by $1/\pi_{\theta_{\rm old}}(a_t \, | \, s_t)$ instead of $1/\pi_{\theta}(a_t \, | \, s_t)$.

\subsection{Baselines} \label{baselines-appendix}
First, we have that
\begin{align*} 
& \underset{a_i}{\int} \dfrac{\partial \pi_{\theta}(a_i \, | \, s_i)}{\partial \theta} da_i  \\
& \,  = \dfrac{\partial}{\partial \theta}\underset{a_i}{\int} \pi_{\theta}(a_i \, | \, s_i) da_i \ \ \ = \ \ \ \dfrac{\partial}{\partial \theta}1 \\
& \, = 0. \stepcounter{equation}\tag{\theequation}\label{app-baseline1}
\end{align*} 
Where the second line requires the interchanging of derivative and integral. This paper will not rigorously justify this interchange, but we note that a formal proof just requires some straightforward assumptions such as $\pi_{\theta}$ being differentiable with respect to $\theta$, and the actions having support which does not depend on $\theta$. 

Let $\beta_{\phi}(\xi_i)$ depend on any $s_1, a_1, r_1, s_2, \ldots, s_i$. Then, 
\begin{align*} 
& \mathe_{(s_1, a_1, r_1)\ldots (s_T, a_T, r_T)} \Big[ \beta_{\phi}(\xi_i) \dfrac{\partial \log \pi_{\theta}(a_i \, | \, s_i)}{\partial \theta} \Big] \\
= & \mathe_{(s_1, a_1, r_1)\ldots (s_i, a_i, r_i)} \Big[ \beta_{\phi}(\xi_i) \dfrac{\partial \log \pi_{\theta}(a_i \, | \, s_i)}{\partial \theta} \Big] \\
 = & \mathe_{(s_1, a_1, r_1)\ldots (s_i)}\Big[ \beta_{\phi}(\xi_i)   \mathe_{a_i} \Big[ \dfrac{\partial \log \pi_{\theta}(a_i \, | \, s_i)}{\partial \theta} \ \Big|\  s_1, a_1, r_1, \ldots, s_i \Big] \Big] \\
 = & \mathe_{(s_1, a_1, r_1)\ldots (s_i)}\Big[ \beta_{\phi}(\xi_i) \big( 0 \big) \Big] \ = \ 0.
\end{align*}
where the third line uses \eqref{app-baseline1}.

\subsection{Q-function Baseline} \label{optimal-2021}
In \cite{beyond-variance}, the authors give a closed form for the optimal state-dependent baseline
\begin{align*} 
\beta_{\phi}(s_i) \, = \, \dfrac{\mathe \bigg[ Q^{\pi}(s_i, a_i) \innerp{\dfrac{\partial \log \pi_{\theta}(a_i \, | \, s_i)}{\partial \theta}, \dfrac{\partial \log \pi_{\theta}(a_i \, | \, s_i)}{\partial \theta}} \bigg]}{\mathe \bigg[ \innerp{\dfrac{\partial \log \pi_{\theta}(a_i \, | \, s_i)}{\partial \theta}, \dfrac{\partial \log \pi_{\theta}(a_i \, | \, s_i)}{\partial \theta}} \bigg]}  \stepcounter{equation}\tag{\theequation}\label{optimal-app-1}
\end{align*}
where $Q^{\pi}(s_i, a_i)$ is the action-value function for a policy $\pi$. The expression is correct for a MDP with $T=1$ (e.g. a multi-arm bandit problem). %However, our analytic example \ref{coin-flips} showed that \eqref{optimal-app-1} is not the true optimal baseline. 

%\subsection*{Toy Examples Supplement}
%In this variant inspired by \cite{beyond-variance}, we start from an initial value of $[\theta_1, \theta_2, \theta_3] = [5, 3, 0]$ (weighted towards the two suboptimal arms). Eventually, all algorithms will converge to the optimal arm, regardless of the choice of baseline. Again we did not observe any significant differences or benefits from having a lower variance, or from adding a positive constant to the baseline values.
%\subsubsection*{Three-Arm Bandit}
%\begin{figure}[H] 
%\centering 
%\includegraphics[ width=\textwidth]{mab variance fig3.png}
%\caption{TODO} 
%\end{figure}

%\nocite{*}
\bibliographystyle{unsrt}
\bibliography{keane_sources}

\begin{thebibliography}{10}

\bibitem{bottou-nocedal}
L\'eon Bottou, Frank~E. Curtis, and Jorge Nocedal.
\newblock Optimization methods for large-scale machine learning.
\newblock {\em SIAM Review}, 60(2):223--311, 2018.

\bibitem{monte-carlo-gradient}
Shakir Mohamed, Mihaela Rosca, Michael Figurnov, and Andriy Mnih.
\newblock Monte carlo gradient estimation in machine learning.
\newblock {\em Journal of Machine Learning Research}, 21(132):1--62, 2020.

\bibitem{a3c}
Volodymyr Mnih, Adria~Puigdomenech Badia, Mehdi Mirza, Alex Graves, Timothy
  Lillicrap, Tim Harley, David Silver, and Koray Kavukcuoglu.
\newblock Asynchronous methods for deep reinforcement learning.
\newblock In Maria~Florina Balcan and Kilian~Q. Weinberger, editors, {\em
  Proceedings of The 33rd International Conference on Machine Learning},
  volume~48 of {\em Proceedings of Machine Learning Research}, pages
  1928--1937, New York, New York, USA, 20--22 Jun 2016. PMLR.

\bibitem{ppo}
John Schulman, Filip Wolski, Prafulla Dhariwal, Alec Radford, and Oleg Klimov.
\newblock Proximal policy optimization algorithms.
\newblock {\em CoRR}, abs/1707.06347, 2017.

\bibitem{Williams}
R.~J. Williams.
\newblock Simple statistical gradient-following algorithms for connectionist
  reinforcement learning.
\newblock {\em Machine Learning}, 8:229--256, 1992.

\bibitem{sutton-88}
Richard~S. Sutton.
\newblock Learning to predict by the methods of temporal differences.
\newblock {\em Mach. Learn.}, 3(1):9–44, aug 1988.

\bibitem{gae}
John Schulman, Philipp Moritz, Sergey Levine, Michael Jordan, and Pieter
  Abbeel.
\newblock High-dimensional continuous control using generalized advantage
  estimation.
\newblock 2015.

\bibitem{stochastic-computation-graphs}
John Schulman, Nicolas Heess, Theophane Weber, and Pieter Abbeel.
\newblock Gradient estimation using stochastic computation graphs.
\newblock In {\em Proceedings of the 28th International Conference on Neural
  Information Processing Systems - Volume 2}, NIPS'15, page 3528–3536,
  Cambridge, MA, USA, 2015. MIT Press.

\bibitem{greensmith-2004}
Evan Greensmith, Peter~L. Bartlett, and Jonathan Baxter.
\newblock Variance reduction techniques for gradient estimates in reinforcement
  learning.
\newblock {\em J. Mach. Learn. Res.}, 5:1471–1530, dec 2004.

\bibitem{credit-assignment-techniques}
T.~Weber, N.~Heess, Lars Buesing, and D.~Silver.
\newblock Credit assignment techniques in stochastic computation graphs.
\newblock In {\em AISTATS}, 2019.

\bibitem{nvil}
Andriy Mnih and Karol Gregor.
\newblock Neural variational inference and learning in belief networks.
\newblock 2014.

\bibitem{nemirovski}
A.~Nemirovski, A.~Juditsky, G.~Lan, and A.~Shapiro.
\newblock Robust stochastic approximation approach to stochastic programming.
\newblock {\em SIAM Journal on Optimization}, 19(4):1574--1609, 2009.

\bibitem{almost-sure-SGD}
Panayotis Mertikopoulos, Nadav Hallak, Ali Kavis, and Volkan Cevher.
\newblock On the almost sure convergence of stochastic gradient descent in
  non-convex problems.
\newblock 2020.

\bibitem{peters-2008}
J.~Peters and S.~Schaal.
\newblock Reinforcement learning of motor skills with policy gradients.
\newblock {\em Neural Networks}, 21(4):682--697, May 2008.

\bibitem{beyond-variance}
Wesley Chung, Valentin Thomas, Marlos~C. Machado, and Nicolas~Le Roux.
\newblock Beyond variance reduction: Understanding the true impact of baselines
  on policy optimization.
\newblock {\em ArXiv}, abs/2008.13773, 2021.

\bibitem{weaver-tao}
Lex Weaver and Nigel Tao.
\newblock The optimal reward baseline for gradient-based reinforcement
  learning.
\newblock 2013.

\bibitem{russo-policy-gradient}
Jalaj Bhandari and Daniel Russo.
\newblock Global optimality guarantees for policy gradient methods.
\newblock 2019.

\bibitem{gym}
Greg Brockman, Vicki Cheung, Ludwig Pettersson, Jonas Schneider, John Schulman,
  Jie Tang, and Wojciech Zaremba.
\newblock Openai gym.
\newblock 2016.

\bibitem{mirage-action}
George Tucker, Surya Bhupatiraju, Shixiang Gu, Richard Turner, Zoubin
  Ghahramani, and Sergey Levine.
\newblock The mirage of action-dependent baselines in reinforcement learning.
\newblock In Jennifer Dy and Andreas Krause, editors, {\em Proceedings of the
  35th International Conference on Machine Learning}, volume~80 of {\em
  Proceedings of Machine Learning Research}, pages 5015--5024. PMLR, 10--15 Jul
  2018.

\end{thebibliography}

\end{document}